\documentclass[11pt,a4paper]{article}
\usepackage[hyperref]{acl2021}
\usepackage{times}

\usepackage{latexsym}
\usepackage{amssymb}
\usepackage{multicol}
\usepackage{multirow}
\usepackage{color}
\usepackage{booktabs}
\usepackage{graphicx}
\usepackage{listings}
\usepackage{xcolor}
\usepackage{url}
\usepackage{array}
\usepackage{pythonhighlight}
\usepackage{microtype}

\aclfinalcopy % Uncomment this line for the final submission
\usepackage[textsize=scriptsize]{todonotes}

\title{CoSQA: 20,000+ Web Queries for Code Search and Question Answering
}

\author{
Junjie Huang$^{1}\thanks{\ \ Work done during internship at Microsoft Research Asia.}$\hspace{0.2em}, 
Duyu Tang$^{2}$, 
Linjun Shou$^{3}$, 
Ming Gong$^{3}$,\\
{\bf Ke Xu$^{1}$,
Daxin Jiang$^{3}$,
Ming Zhou$^{2}$,
Nan Duan$^{2}$}\hspace{0.5em}\\

$^{1}$Beihang University \quad
$^{2}$Microsoft Research Asia \quad
$^{3}$Microsoft STC Asia  \\
{\tt $^{1}$huangjunjie@buaa.edu.cn, kexu@nlsde.buaa.edu.cn}\\
{\tt $^{2,3}$\{dutang,lisho,migon,djiang,nanduan\}@microsoft.com  }
}

\date{}

\begin{document}
\maketitle
\begin{abstract}
Finding codes given natural language query is beneficial to the productivity of software developers. 
Future progress towards better semantic matching between query and code requires richer supervised training resources.
To remedy this, we introduce the CoSQA dataset. It includes 20,604 labels for pairs of natural language queries and codes, each annotated by at least 3 human annotators.
% To better leverage the data, we
We further introduce a contrastive learning method dubbed CoCLR to enhance query-code matching, which 
% which further leverages CoSQA and 
% extends SimCLR by considering a pair of data as the input 
works as a data augmenter to bring more artificially generated training instances.
We show that evaluated on CodeXGLUE with the same CodeBERT model, training on CoSQA improves the accuracy of code question answering by 5.1\%, and incorporating CoCLR brings a further improvement of 10.5\%.
% We present . crowdworkers
% Additionally, we propose a new model for stance detection that implicitly captures relationships between topics using generalized topic representations and show that this model improves performance on a number of challenging linguistic phenomena
\footnote{The CoSQA data and leaderboard are available at \href{https://github.com/microsoft/CodeXGLUE/tree/main/Text-Code/NL-code-search-WebQuery}{https://github.com/microsoft/CodeXGLUE/tree/main/Text-Code/NL-code-search-WebQuery}. The code is available at \href{https://github.com/Jun-jie-Huang/CoCLR}{https://github.com/Jun-jie-Huang/CoCLR}}.
\end{abstract}

\section{Introduction}

% \textcolor{blue}{
With the growing population of software developers, natural language code search, which improves the productivity of the development process via retrieving semantically relevant code given natural language queries, is increasingly important in both communities of software engineering and natural language processing \cite{miltos2018survey, Liu2020OpportunitiesAC}. 
The key challenge is how to effectively measure the semantic similarity between a natural language query and a code. 
% }

\begin{figure}
    \includegraphics[width=7.8cm]{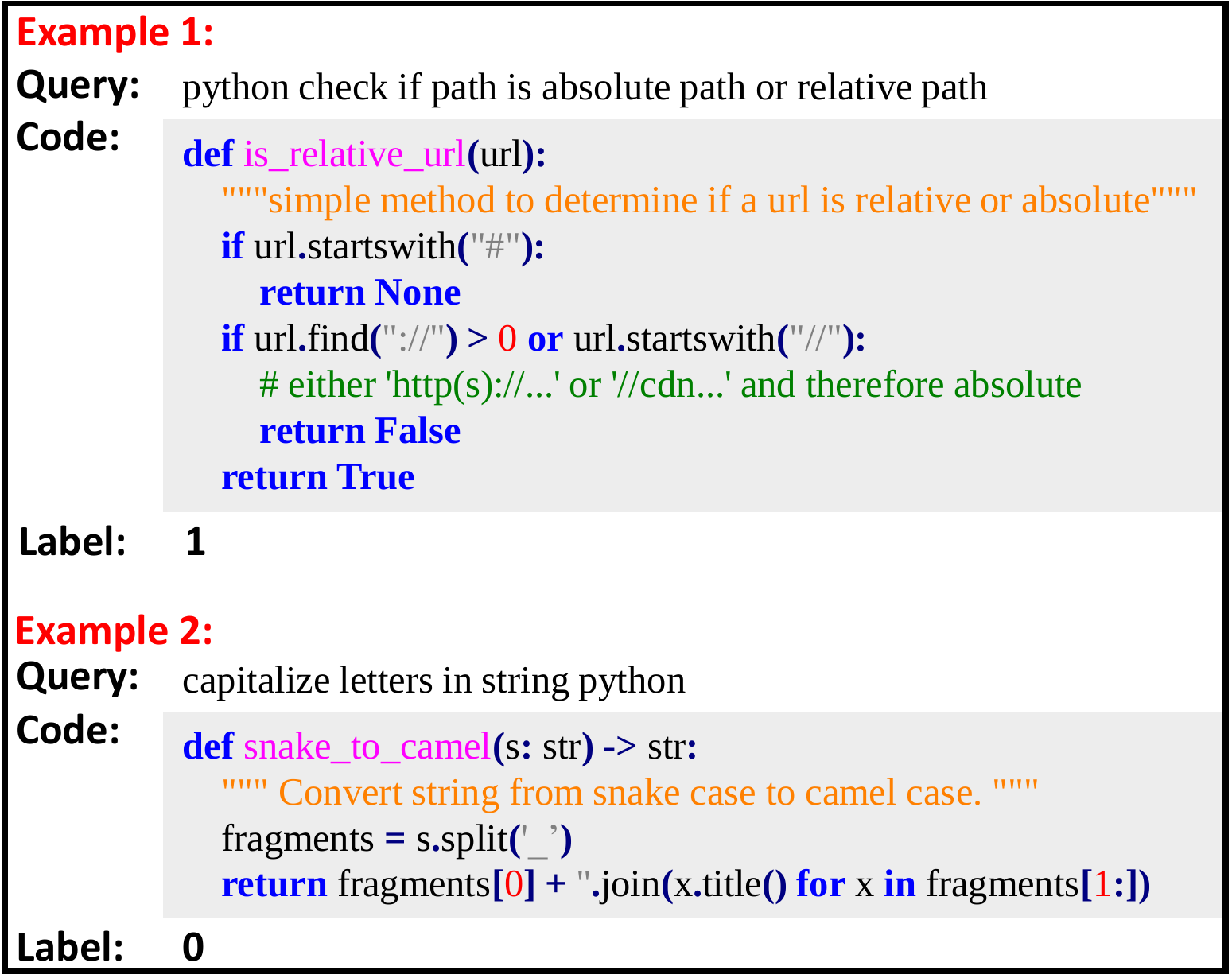}
    \caption{Two examples in CoSQA. A pair of a web query and a Python function with documentation is annotated with ``1" or ``0", representing whether the code answers the query or not.}
    \label{fig:example}
\end{figure}

% \textcolor{blue}{
There are recent attempts to utilize deep neural networks \cite{gu2018deepcs, yao2019mulmodalatt, Feng2020CodeBERTAP}, which embed query and code as dense vectors to perform semantic matching in a unified vector space.
However, these models are mostly trained on pseudo datasets in which a natural language query is either the documentation of a function or a tedious question from Stack Overflow. 
Such pseudo queries do not reflect the distribution of real user queries that are frequently issued in search engines. 
To the best of our knowledge, datasets that contain real user web queries include \citet{Lv2015CodeHowEC}, CodeSearchNet Challenge \cite{husain2019codesearchnet}, and CodeXGLUE
\footnote{\href{https://github.com/microsoft/CodeXGLUE}{https://github.com/microsoft/CodeXGLUE}}
\cite{codexglue}. These three datasets only have 34, 99, and 1,046 queries, respectively, for model testing. The area lacks a dataset with a large amount of real user queries to support the learning of statistical models like deep neural networks for matching the semantics between natural language web query and code.
% }
% It is commonly accepted that However, to the best of our knowledge, the 

\begin{table*}[htbp]
\small
  \centering
    % \resizebox{1.0\textwidth}{!}{
    % \begin{tabular}{p{6.5em}p{4.6em}p{5.875em}p{3.6em}p{6.2em}p{5.6em}p{6.1em}p{6.065em}}
    \begin{tabular}{llllc}
    \toprule
    Dataset  & Size  & Natural Language & Code & human-annotated ? \\
    \midrule
    CodeSearchNet \cite{husain2019codesearchnet} & 2.3M & Documentation & Function & No \\
    \citet{gu2018deepcs} & 18.2M & Documentation & Function & No \\
    \citet{sennrich2017codedoc} & 150.4K & Documentation & Function & No \\
    StaQC (manual) \cite{yao2018staqc}  & 8.5K & Stack Overflow question & Code block  & Yes \\
    StaQC (auto) \cite{yao2018staqc} & 268K & Stack Overflow question & Code block  & No \\
    CoNaLa (manual) \cite{yin2018mining} & 2.9K  & Stack Overflow question  & Statements &  Yes  \\
    CoNaLa (auto) \cite{yin2018mining} & 598.2K & Stack Overflow question & Statements & No \\
    SO-DS \cite{Heyman2020SODS}  & 12.1K & Stack Overflow question & Code block  & No \\
    % \citet{iyer2016summarizing} & 98.3K & Stack Overflow question & Code block  & No \\
    \citet{Nie2016QueryEB}   & 312.9K & Stack Overflow question & Code block  & No \\
    \citet{Li2019NeuralCS} & 287 & Stack Overflow question & Function & Yes \\
    \citet{yan2020benchcs} & 52 & Stack Overflow question & Function & Yes \\
    % \citet{Oda2015LearningTG}   & 18.8K &  Human summarization & Statements & Yes \\
    \midrule
    \citet{Lv2015CodeHowEC}  &  34 & Web query & Function & Yes \\  
    CodeSearchNet \cite{husain2019codesearchnet} & 99 & Web query & Function & Yes \\
    CodeXGLUE WebQueryTest \footnotemark[2] & 1K & Web query & Function & Yes \\
    % \midrule
    CoSQA (ours) & 20.6K & Web query & Function & Yes \\
    \bottomrule
    \end{tabular}%
    % }
  \caption{Overview of existing datasets on code search and code question answering. 
%   Check mark indicate datasets with human annotations.
    Some datasets containing both unlabelled data and labelled data are listed in separate lines.}
  \label{tab:query-code-data}%
\end{table*}%

To address the aforementioned problems, we introduce CoSQA, a dataset with 20,604 pairs of web queries and code for \textbf{Co}de \textbf{S}earch and \textbf{Q}uestion \textbf{A}nswering, each with a label indicating whether the code can answer the query or not. The queries come from the search logs of the Microsoft Bing search engine, and the code is a function from GitHub\footnote{We study on Python in this work, and we plan to extend to more programming languages in the future.}. 
To scale up the annotation process on such a  professional task, we elaborately curate potential positive candidate pairs and perform large scale annotation where each pair is annotated by at least three crowd-sourcing workers. 
% Results on our pilot annotation demonstrate the effectiveness of the annotation pipeline. 
Furthermore, to better leverage the CoSQA dataset for query-code matching,
% \junjie{revise} considering the characteristics that web queries are often brief and not necessarily grammar correct, 
we propose a code contrastive learning method (CoCLR) 
% that extends SimCLR framework \cite{chen2020simclr} 
to produce more artificially generated instances for training.
% which includes paired negative examples within mini-batch and augmented pseudo positive examples augmented
% \junjie{to produce more artificially generated training instances}

We perform experiments on the task of query-code matching on two tasks: code question answering and code search.
% For code question answering, the performance on CodeXGLUE WebQueryTest improves 5.1\% after utilizing CoSQA dataset, and further boosts 10.5\% by incorporating our CoCLR method. 
On code question answering, we find that the performance of the same CodeBERT model improves 5.1\% after training on the CoSQA dataset, and further boosts 10.5\% after incorporating our CoCLR method.
Moreover, experiments on code search also demonstrate similar results. 
% Finally, we conduct detailed analyses to explore the factors influencing code search results, aiming to point out the characteristics and difficulties of the query-code matching tasks.

\section{Related Work}
% In this section, we begin with a survey of existing query-code matching datasets. Table \ref{tab:query-code-data} illustrates an overview of these datasets. We also introduce models for natural language code search, a representative task of query-code matching. 
In this part, we describe existing datasets and methods on code search and code question answering. 

\subsection{Datasets}
A number of open-sourced datasets with a large amount of text-code pairs have been proposed for the purposes of code search \cite{husain2019codesearchnet, gu2018deepcs, Nie2016QueryEB}
% code summarization \cite{iyer2016summarizing, sennrich2017codedoc, Oda2015LearningTG}, 
and code question answering \cite{yao2018staqc, yin2018mining, Heyman2020SODS}.
% which can be applied for training code search models. 
There are also high-quality but small scale testing sets curated for code search evaluation \cite{Li2019NeuralCS, yan2020benchcs, Lv2015CodeHowEC}. 
\citet{husain2019codesearchnet}, \citet{gu2018deepcs} and \citet{sennrich2017codedoc} collect large-scale unlabelled text-code pairs by leveraging human-leaved comments in code functions from GitHub. \citet{yao2018staqc} and \citet{yin2018mining} automatically mine massive code answers for Stack Overflow questions with a model trained on a human-annotated dataset.
% \citet{iyer2016summarizing} and 
\citet{Nie2016QueryEB} extract the Stack Overflow questions and answers with most likes to form text-code pairs.
% \citet{Oda2015LearningTG} propose a pseudo-code summarization dataset by writing summaries for each line of code.
Among all text-code datasets, only those in \citet{Lv2015CodeHowEC}, CodeSearchNet Challenge \cite{husain2019codesearchnet} and CodeXGLUE\footnotemark[2] contain real user web queries, but they only have 34, 99 and 1,046 queries for testing and do not support training data-driven models. Table \ref{tab:query-code-data} illustrates an overview of these datasets.

% Besides natural language code search, \citet{rohan2020contextcs} proposed contextual code search where the query is previous code context.

\subsection{Code Search Models}
% mainly lies in two directions:
Models for code search mainly can be divided into two categories: information retrieval based models and deep learning based models. Information retrieval based models match keywords in the query with code sequence \cite{Bajracharya2006SourcererAS, Liu2020SimplifyingDM}. Keyword extension by query expansion and reformulation is an effective way to enhance the performance \cite{Lv2015CodeHowEC, Lu2015QueryEV, Nie2016QueryEB, Rahman2019AutomaticQR, Rahman2019SupportingCS}.  
deep learning based models encode query and code into vectors and utilize vector similarities as the metric to retrieve code  \cite{Sachdev2018RetrievalOS, Ye2016FromWE, gu2018deepcs, Cambronero2019WhenDL, yao2019coacor,
liu2019queryexpand, Feng2020CodeBERTAP, zhao2020adversarialcodesearch}. There are also ways to exploit code structures to learn better representations for code search \cite{yao2019mulmodalatt, haldar2020multi, Guo2020GraphCodeBERTPC}.

% \begin{itemize}
%     \item In order to determine a more realistic distribution of NL queries, we collect web queries from a commercial search engine and perform manual analysis. In particular, we find that a large number of web queries can not be answered only by code. Therefore, we design heuristic rules to filter out these queries (Section \ref{sec:searchintent}).

\section{CoSQA Dataset}
In this section, we introduce the construction of the CoSQA dataset. We study Python in this work, and we plan to extend to more programming languages in the future.
Each instance in CoSQA is a pair of natural language query and code, which is annotated with ``1" or ``0" to indicate whether the code can answer the query.
We first describe how to curate web queries, obtain code functions, and get candidate query-code pairs. 
After that, we present the annotation guidelines and statistics.
% For each web query, a function is annotated with a label to indicate whether the code can answer the query. 
% To collect data, we first acquire candidate pairs from scratch in three stages: curating web queries, obtaining code functions, and pairing them to form candidate pairs. 
% Then we crowdsource the annotation task to label the candidates.

% Here we select Python as our target programming language since it is popular worldwide and easy for machine learning practitioners to understand, but we are looking forward to extend annotation to other languages.

\subsection{Data Collection}
% To pair possible code function answers for web queries, we first collect  
% First, we collect user web search queries with potential code search intent. Next we automatically generate possible code functions to answer the queries by a powerful code search model.  Finally we obtain pairs of query and code function ready for annotation.

\paragraph*{Query Curation}\label{sec:searchintent}
% \subsubsection*{Filtering Query}\label{sec:searchintent}
% \junjie{ask linjun for details}
% 向linjun laoshi求证这些query的获取方法, 以及一些处理过程
We use the search logs from the Microsoft Bing search engine as the source of queries. Queries without the keyword ``\textit{python}" are removed.
% First, we collect real user search queries from a commercial search engine and only keep queries containing the keyword of \textit{python}. 
Based on our observation and previous work \cite{yao2018staqc, yan2020benchcs}, there are seven basic categories of code-related web queries, including: (1) code searching, (2) debugging, (3) conceptual queries, (4) tools usage, (5) programming knowledge, (6) vague queries and (7) others. 
% Table \ref{tab:code-search-intent} lists an example and some keywords for the seven categories. % definition这个加不加呢 还是是用feature
Basically, queries in (2)-(7) categories are not likely to be answered only by a code function, since they may need abstract and general explanations in natural language. Therefore, we only target the first category of web queries that have code searching intent, i.e., queries that can be answered by a piece of code.

To filter out queries without code searching intent, we manually design heuristic rules based on exact keyword matching. For example, queries with the word of \textit{benefit} or \textit{difference} are likely to seek a conceptual comparison rather than a code function, so we remove all queries with such keywords. Based on the observations, we manually collect more than 100 keywords in total. Table \ref{tab:code-search-intent} displays a part of selected keywords used for removing unqualified queries and more details can be found in Appendix \ref{appen:query-filter-rules}. To evaluate the query filtering algorithm, we construct a human-annotated testset. We invite three experienced python programmers to label 250 randomly sampled web queries with a binary label of having/not having searching intent. Then we evaluate the accuracy of intent predictions given keyword-based rules and those given by humans. We find the F1 score achieves 67.65, and the accuracy is up to 82.40. This demonstrates the remarkable effectiveness of our rule-based query filtering algorithm.
% We evaluate our heuristic rules on a randomly sampled python web query testset which contains XX \todo{stat.} queries along with binary annotations of intent from experienced python programmers.
% Finally, XX\% \todo{stat.} of python web queries are remained after intent filtering process. 

\begin{table}[t]
\small
  \centering
%   \resizebox{1.0\textwidth}{!}{
    % \begin{tabular}{p{3cm}p{8cm}p{2cm}}
    \begin{tabular}{m{1.5cm}<{\centering}m{5.4cm}}
    
    \toprule
    Categories & Some Keywords   \\
    \midrule
    Debugging & 
            exception, index out of, ignore, stderr, \dots
            % try \dots except, debug, no such file or directory, warning, 
        \\ 
    \specialrule{0em}{0pt}{4pt}
    Conceptual Queries & 
            vs, versus, difference, advantage, benefit, drawback, how many, what if, why, \dots
            % interpret, understand, 
            % cannot, can’t, couldn’t, could not, 
            % how many, how much, too much, too many, more, less,
            % what if, what happens, what is, what are,
            % when, where, which, why, reason,
            % how do \dots work, how \dots works, how does \dots work,
            % need, require, wait, turn \dots on/off, turning \dots on/off, 
        \\ 
    \specialrule{0em}{0pt}{4pt}
    Programming Knowledge & 
            tutorial, advice, argument, suggestion, statement, declaration, operator, \dots
            % tutorial, advice, course, proposal, discuss, suggestion,
            % parameter, argument, statement, class, import,
            % inherit, operator, override, decorator, descriptor,
            % declare, declaration
        \\ 
    \specialrule{0em}{0pt}{4pt}
    Tools Usage & 
            console, terminal, open python, studio, ide,
            ipython, jupyter, vscode, vim, \dots
            % console, terminal, open python, studio, ide,
            % ipython, jupyter, notepad, notebook, vim,
            % Pycharm, vscode, eclipse, sublime, emacs, utm,
            % komodo, pyscripter, eric, c\#, access control,
            % pip, install, library, module, launch, version,
            % ip address, ipv, get \dots ip, check \dots ip, valid \dots ip,
        \\ 
    \specialrule{0em}{0pt}{4pt}
    Others & 
           unicode, python command, ``@", ``()", \dots
            % ``()", ``.", ``\_", ``:", ``@", ``=", ``\textgreater", ``\textless", ``-"
        \\
    
    \bottomrule
    \end{tabular}%
%   }
  \caption{Selected keywords for our heuristic rules to filter out web queries without code search intent in five categories. Vague queries are morphologically variable, so we ignore this category.} 
  \label{tab:code-search-intent}%
\end{table}%

% Then we filter out queries without code search intent.
% We only target at queries with code search intent in XXXXX, i.e. queries can be answered by a piece of code, because answers to other queries without code search intent are not very likely to be a code function. 
% Following previous work \cite{yao2018staqc, yan2020benchcs}, we conclude six basic categories of queries without code search intent, including: (1) debugging, (2) conceptual, (3) tools usage, (4) programming knowledge, (5) unclear intents, and (6) others. 
% Queries in the first four categories can be recognized and filtered out by some simple rules of keyword overlapping. 
% Table \ref{table-code-search-intent} lists the definition, an example and some keywords for each of the six categories.
% Note that our keyword overlapping-based filtering method can not guarantee removing all queries without code search intent. With intent filtering process, we expect to increase the proportion of positive query-code pairs in the annotation stage. Finally we acquire NN-NN queries.

% \subsubsection*{Collecting Code}
\paragraph{Code Collection}
The selection of code format is another important issue in constructing query-code matching dataset, which includes a statement \cite{yin2018mining}, a code snippet/block \cite{yao2018staqc}, a function \cite{husain2019codesearchnet}, etc. 
In CoSQA, we simplify the task and adopt a compete Python function with paired documentation to be the answer to the query for the following reasons.
First, it is complete and independent in functionality which may be more prone to answering a query. 
Second, it is syntactically correct and formally consistent which enables parsing syntax structures for advanced query-code matching. 
Additionally, a complete code function is often accompanied with documentation wrote by programmers to help understand its functionality and usage, which is beneficial for query-code matching (see Section \ref{sec:code-component} for more details). 
% \junjie{any better reason why we add doc?} Therefore, we use a function as the code in CoSQA.

% We define a Python function to be the form of answer for code search since it is complete and can be used to parse structures and syntax.
% 这点合适吗
% We also include the original paired documentation in the functions to encourage the exploration of the roles of documentation.
We take the CodeSearchNet Corpus \cite{husain2019codesearchnet} as the source for code functions, which is a large-scale open-sourced code corpus allowing modification and redistribution. The corpus contains 2.3 million functions with documentation and 4.1 million functions without documentation from public GitHub repositories spanning six programming languages (Go, Java, JavaScript, PHP, Python, and Ruby). % 需要再这后面再说我们为什么选CodeSearchNet吗
In CoSQA, we only keep complete Python functions with documentation and remove those with non-English documentation or special tokens (e.g. $``\langle img ... \rangle"$ or $``http://"$). 
% We remove functions without documentation or with non-English documentation and finally collect a set of NNN-N functions.
% Finally we collect XXXXX valid Python functions.

\begin{table*}[t]
\small
  \centering
  \resizebox{1.0\textwidth}{!}{
    % \begin{tabular}{p{3cm}p{8cm}p{2cm}}
    \begin{tabular}{m{2.8cm}<{\centering}m{7.8cm}<{\centering}m{4.3cm}}
    
    \toprule
    Query & Code  & Explanations  \\
    \midrule
    (1) boolean function to check if variable is a string python & 
            \includegraphics[width=8cm]{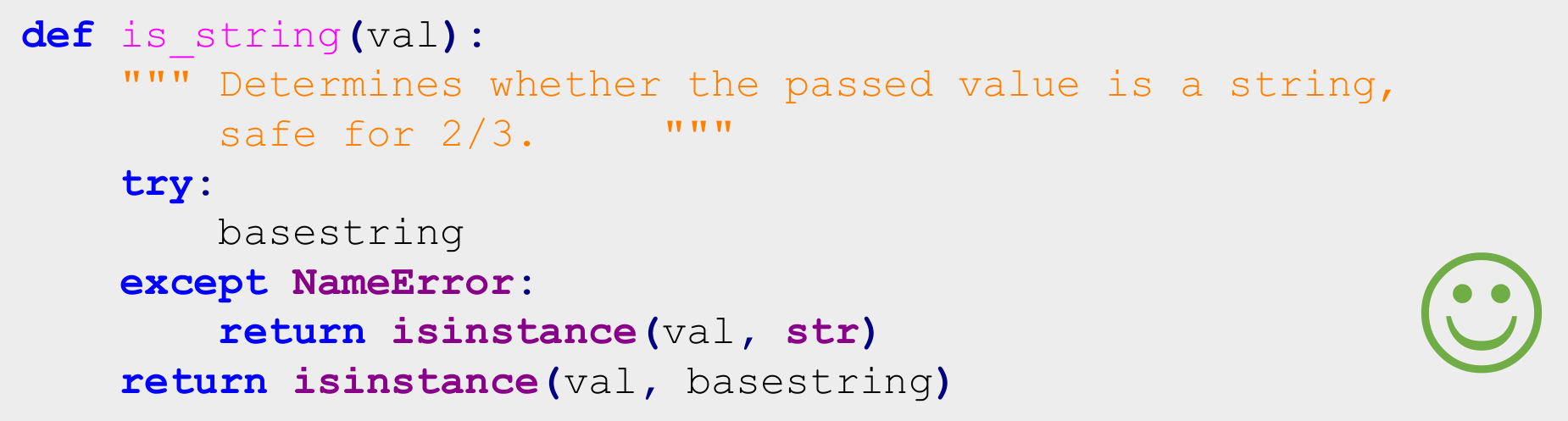} & 
            Code can fully satisfy the demand of the query. Therefore the code is a correct answer.
             \\
    \hline
    (2) python check if argument is list & 
            \includegraphics[width=8cm]{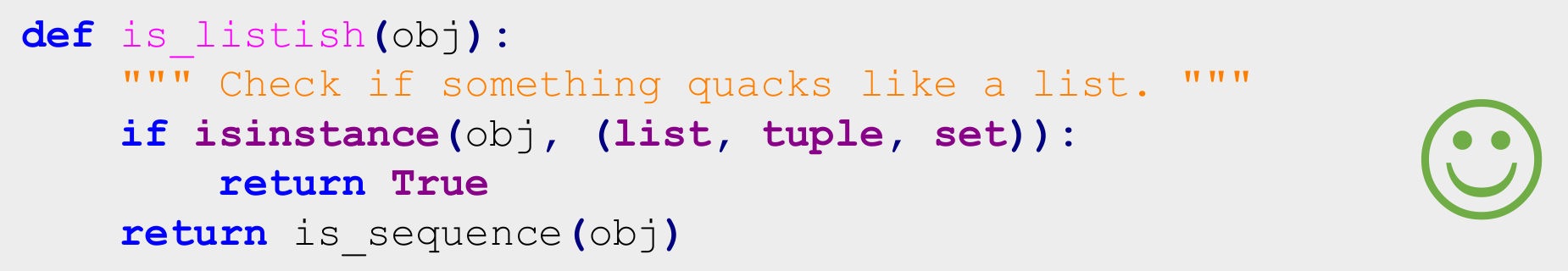} & 
            Code meets the demand of checking \textit{list} type, and the \textit{tuple} and \textit{set} types, which exceeds query's demand. It is a correct answer.
             \\
    \hline
   (3). python measure distance between 2 points & 
            \includegraphics[width=8cm]{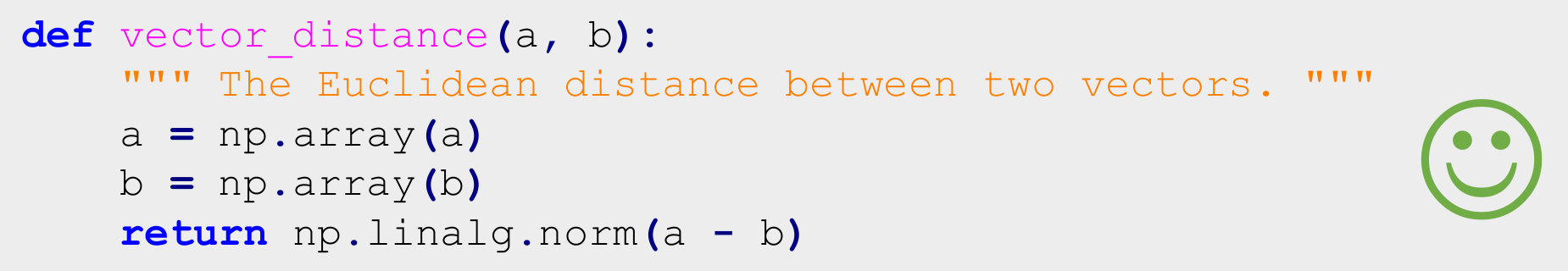}  & Code computes Euclidean distance, which is one category of vector distances. So it is correct.
            \\
    (4) python measure distance between 2 points & 
            \includegraphics[width=8cm]{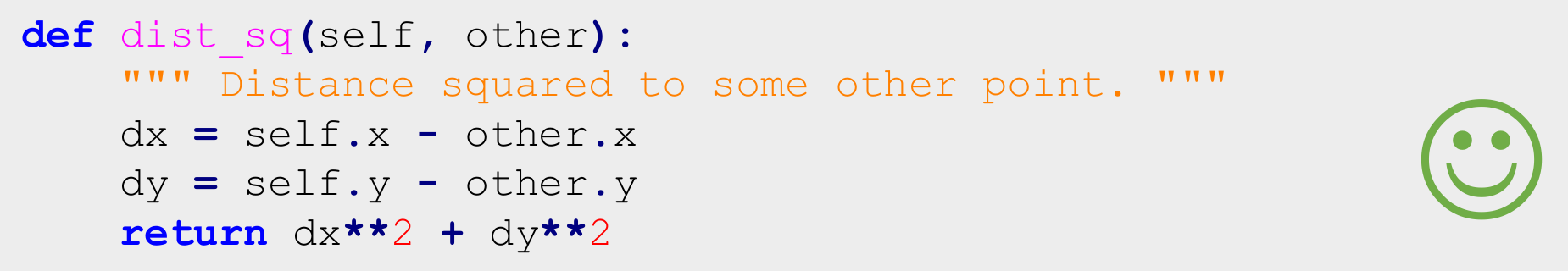}  & 
            Code computes square distance, which is another category of vector distances. 
            \\
    \hline
    (5) read write in the same file python & 
            \includegraphics[width=8cm]{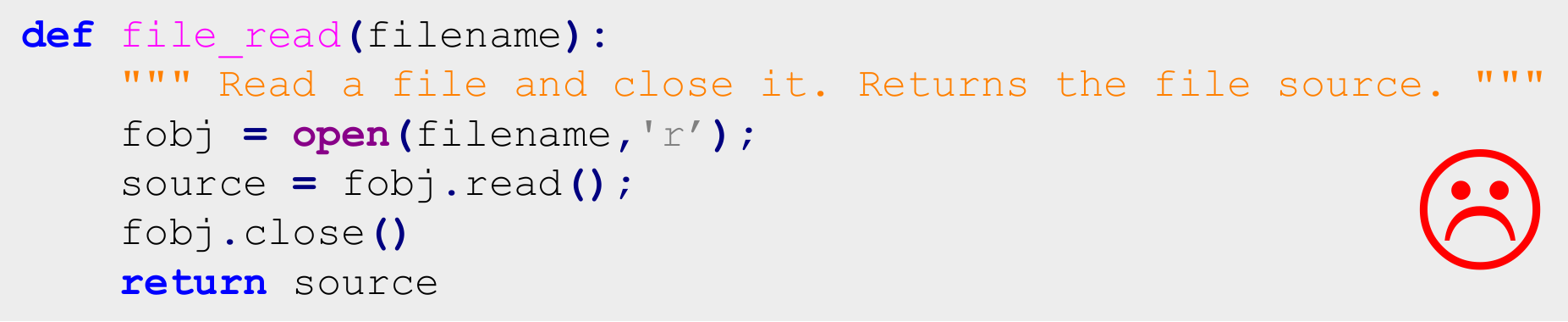}   & 
            Query asks for reading and writing, but code only implements reading. The code satisfies 50\% of the demands and is not a correct answer.
            \\
    (6) python get the value in the list starting with the str &  
            \includegraphics[width=8cm]{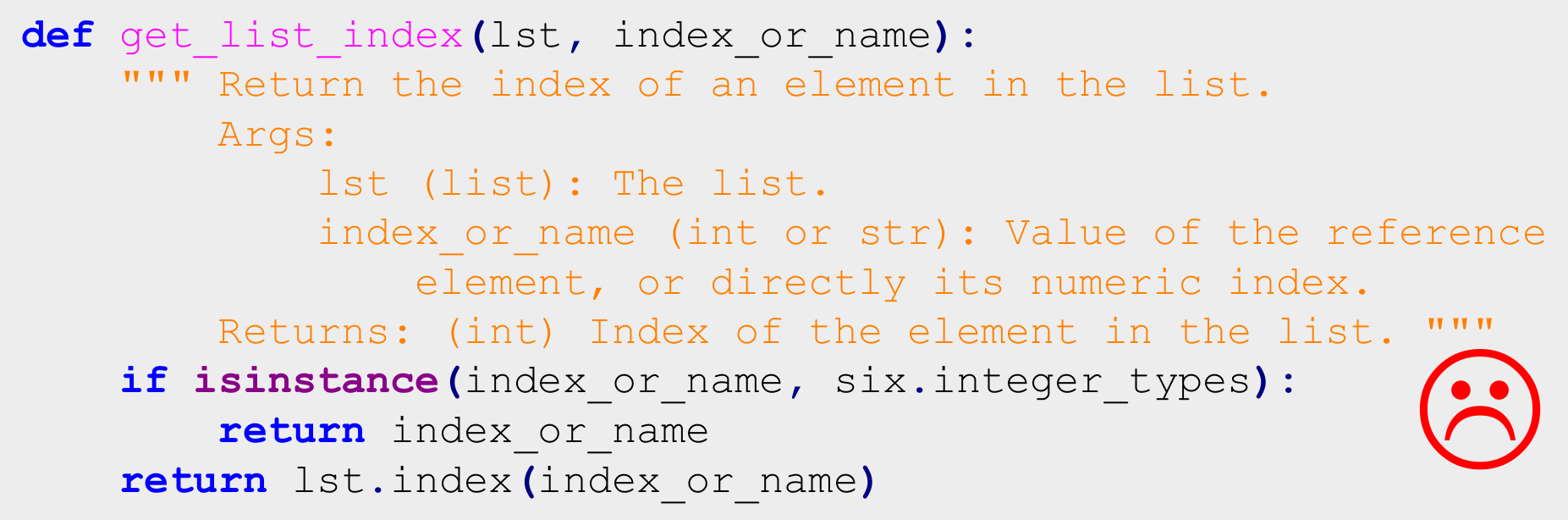}  & 
            The query is looking for an element in the list that starts with a specific \textit{str}, but the code does not have the function of starting with the \textit{str}, and it returns index instead of value. There are two unsatisfied areas, which is less than 50\%.
            \\
    \hline
    (7) python check if something is an array & 
            \includegraphics[width=8cm]{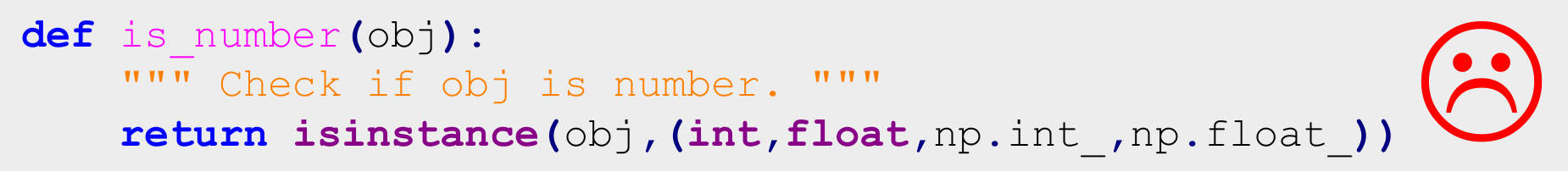}  & 
            A small part of code is relevent to the query but is can not answer.
            \\
    
    \bottomrule
    \end{tabular}%
  }
%   \caption{Examples of query-code pairs for each situation. Notes are also provided to explain why we make the judgement.}   
  \caption{Examples and explanations of query-code pairs for correct and incorrect answers.} 
  \label{tab:example-partial-answer}%
\end{table*}%

% \subsubsection*{Generating Candidate text-code Pairs}
\paragraph{Candidate Query-code Pairs}
Obviously, it is not possible to annotate all query-code pairs. % 笛卡尔积
To improve efficiency, we wipe off low-confidence instances before annotation. 
Specifically, we employ a CodeBERT-based matching model \cite{Feng2020CodeBERTAP} to retrieve high-confidence codes for every query. The CodeBERT encoder is fine-tuned on 148K automated-minded Python Stack Overflow question-code pairs (StaQC) \cite{yao2018staqc} with the default parameters. A cosine similarity score on the pooled $[CLS]$ embeddings of query and code is computed to measure the relatedness. To guarantee the quality of candidates, we automatically remove low-quality query-code pairs according to the following evaluation metrics.
\begin{itemize}
    \item To ensure the code may answer the query, we only keep the code with the highest similarity to the query and remove the pairs with a similarity below 0.5. 
    % The remaining codes are considered unable to answer the query.
    % \item To enlarge the proportion of positive pairs, we manually set a similarity threshold. The candidate with a similarity score less than 0.5 is considered to be of low-quality.
    \item To increase the code diversity and control the code frequency, we restrict the maximum occurrence of each code to be 10.
\end{itemize}
% \begin{itemize}
%     \item To ensure the code may answer the query, we only keep the code with the highest similarity to the query. The remaining codes are considered unable to answer the query.
%     \item To enlarge the proportion of positive pairs, we manually set a similarity threshold. The candidate with a similarity score less than 0.5 is considered to be of low-quality.
%     \item To increase the code diversity and control the code frequency, we restrict the maximum occurrence of each code to be 10.
% \end{itemize}

% Finally, we obtained 27NNN pairs of web search queries and codes over 8xxx different code functions.

% For each query-code pair, we manually set a 
% Obviously, it is not possible to annotate all query-code pairs. To shrink the candidates for each query to be annotated, we retrieve the top 1 candidate function per query with the similarity score computed by a CodeBERT-based code search model, which is trained on 148K automated-minded Python Stack Overflow question-code pairs (StaQC) \cite{yao2018staqc} with the default parameters provided by \citet{Feng2020CodeBERTAP}.
% We also discard query-code pairs whose similarity is below a manually selected threshold since they are less likely to be a positive example. Finally, we collect N-NNN pairs of query-code pairs for annotation.

\subsection{Data Annotation}
Annotating such a domain-specific dataset is difficult since it requires the knowledge of Python. Even experienced programmers do not necessarily understand all code snippets. To ensure the feasibility and control annotation quality, we design comprehensive annotation guidelines and take a two-step annotation procedure.
% where annotators and data size are different at each stage. 

% \subsubsection*{Annotation Guidelines}
\paragraph{Annotation Guidelines}
Our annotation guideline is developed through several pilots and further updated with hard cases as the annotation progresses.
% \footnote{We will release the annotation guideline upon acceptance.}
% \footnote{The final version of our annotation guideline is available at: }
Annotation participants are asked to make a two-step judgment for each instance: intent annotation and answer annotation.

In the first step of \textit{intent annotation}, annotators are asked to judge whether the query has the intent to search for a code. They will skip the second step if the query is without code search intent. 
As shown in Section \ref{sec:searchintent}, \textit{vague queries} are hard to be filtered out by our heuristic intent filtering algorithm. 
Therefore, it is necessary to take this step to remove such queries so that we can focus more on the matching between query and code rather than query discrimination.
% Therefore, it is necessary to take this step in order to focus on query-code matching. 
% For better understanding, we provide descriptions and examples of the seven categories of queries to annotators and emphasize the \textit{vague queries} category since they are less prone to be filtered out by our heuristic rules in Section \ref{sec:searchintent}.

In the second step of \textit{answer annotation}, annotators are asked to judge whether the code can answer the query. They should label the instance with ``1" if the code is a \textit{correct answer}; otherwise, it is labeled ``0". In this step, judgment should be made after comprehensively considering the relevance between query with documentation, query with function header, and query with function body.

During annotation, it is often the case that a code function can completely answer the query, which means that the code can satisfy all the demands in the query and it is a correct answer. (Case (1) in Table \ref{tab:example-partial-answer}.) But more often, the code can not completely answer the query. It may exceed, partially meet or even totally dissatisfy the demands of the query. 
Therefore we divide such situations into four categories and give explanations and examples (Table \ref{tab:example-partial-answer}) for each category:
\begin{itemize}
    \item If code can answer the query and even exceed the demand of the query, it is a correct answer. (Case (2) in Table \ref{tab:example-partial-answer}.)
    \item If code can meet a certain category of the query demands, it is also a correct answer. (Case (3) and Case (4) in Table \ref{tab:example-partial-answer}.)
    \item If code satisfies no more than 50\% of the query demands, the code can not correctly answer the query. (Case (5) and Case (6) in Table \ref{tab:example-partial-answer}.)
    \item If a small part of the code is relevant to the query, the code can not be a correct answer. (Case (7) in Table \ref{tab:example-partial-answer}.)
\end{itemize}

% Since cases are common that code can not completely answer the query, we restrict the answer criteria that only code can completely answer the query, or the code can partially answer more than 50\% of the query, the instance can be labeled as a positive example. Examples of completely answer and partially answer can be found in Table \ref{tab:example-partial-answer}.

% \subsubsection*{Annotation}
\paragraph{Annotation}
We ask more than 100 participants, who all have a good grasp of programming knowledge, to judge the instances according to the annotation guideline. Participants are provided with the full guidelines and allowed to discuss and search on the internet during annotation.
% Each instance is first judged by four annotators. Those with equal votes are passed to the second round of annotation, where two more judgments are made by three more annotators.
When annotation is finished, each query-code pair has been annotated by at least three participants. We remove the pairs whose inter-annotator
agreement (IAA) is poor, where Krippendorff’s alpha
coefficient \cite{Krippendorff1980KrippendorffKC} is used to measure IAA. We also remove pairs with no-code-search-intent queries. 
Finally, 20,604 labels for pairs of web query and code are retained, and their average Krippendorff’s alpha coefficient is 0.63.
Table \ref{tab:cosqa-statistics} shows the statistics of CoSQA. 

\begin{table}[h]
\small
  \centering
  
    % \begin{tabular}{c|c}
    % \hline
    % \# of pairs & \% of pairs with label ``1"  \\
    % \hline
    % 20,604 & 51.45\% \\
    % \hline
    % \end{tabular}%
   
    \begin{tabular}{c|c|c|c}
    \hline
    & \# of query & avg. length &  \# of tokens \\
    \hline
    query & 20,604 
          & 6.60 
          & 6,784 \\
    \hline
    code  & 6,267 
          & 71.51 
          &  28,254 \\
    \hline
    \end{tabular}%
  \caption{Statistics of our CoSQA dataset.}
  \label{tab:cosqa-statistics}%
\end{table}%

\section{Tasks}

% Text-code matching aims to measure the semantic relatedness between natural language and code. 
Based on our CoSQA dataset, we explore two tasks to study the problem of query-code matching: code search and code question answering.

The first task is natural language code search, where we formulate it as a text retrieval problem. 
Given a query $q_{i}$ and a collection of codes $C = \{c_{1},  \dots, c_{H}\}$ as the input, the task is to find the most possible code answer $c^{*}$. 
% Given a query $q_{i}$ from a query set $Q = \{q_{1}, \dots, q_{K}\}$, the task is to find the most possible code answer $c^*$ from a fixed collection of codes $C = \{c_{1},  \dots, c_{H}\}$. 
The task is evaluated by Mean Reciprocal Rank (MRR).

The second task is code question answering, where we formulate it as a binary classification problem. 
Given a natural language query $q$ and a code sequence $c$ as the input, the task of code question answering predicts a label of ``1" or ``0" to indicate whether code $c$ answers query $q$ or not. 
% Predict ``1" if $c$ is a satisfying answer, otherwise ``0". 
The task is evaluated by accuracy score.

\section{Methodology}
% In this section, we introduce our Code Contrastive Learning method (CoCLR). 
In this section, we first describe the model for query-code matching and then present our code contrastive learning method (CoCLR) to augment more training instances. 
% architecture, where a siamese CodeBERT classifier is applied to measure the relation of query and code.
% Then we incorporate contrastive learning strategy into the model, i.e. CoCLR, to better leverage the data to train a query-code matching model. 

% Figure \ref{fig:coclr-model} illustrates the framework of siamese CodeBERT and our CoCLR method.

% \begin{figure}
%     \includegraphics[width=7.8cm]{figure/CoCLR_model.pdf}
%     \caption{The framework of CoCLR method.}
%     \label{fig:coclr-model}
% \end{figure}

\begin{figure*}
    \includegraphics[width=\textwidth]{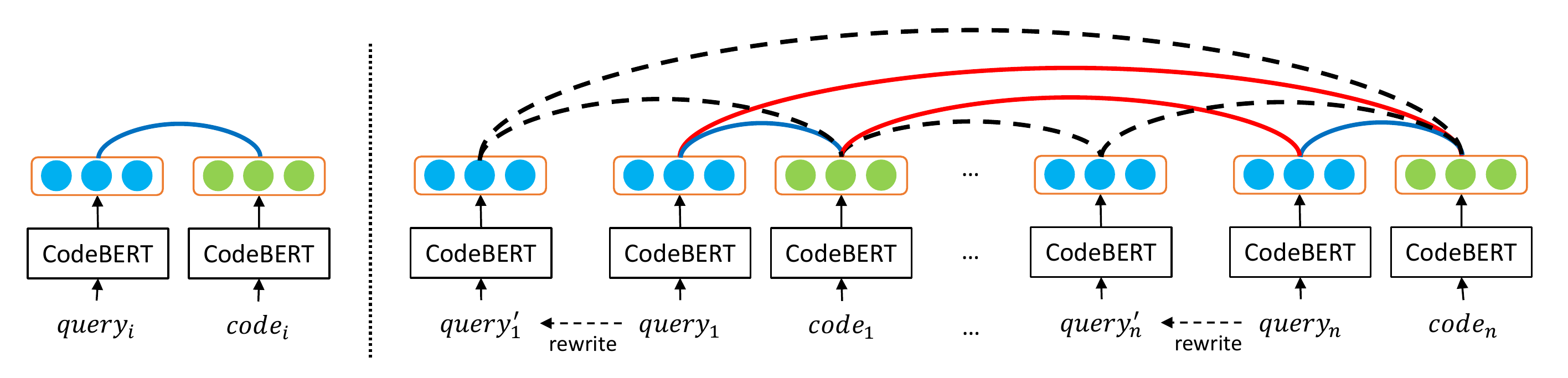}
    \caption{The frameworks of the siamese network with CodeBERT (left) and our CoCLR method (right).
    % The blue line between query and code embeddings denotes the similarity for original instance. The red lines and dashed lines denote similarity for negative examples and augmented pseudo positve examples, respectively.
    The blue line denotes the original training example. The red lines and dashed lines denote the augmented examples with in-batch augmentation and query-rewritten augmentation, respectively.
    }
    \label{fig:coclr-model}
\end{figure*}

\subsection{Siamese Network with CodeBERT} 
The base model we use in this work is 
% Our Code Contrastive Learning (CoCLR) method uses
a siamese network, which is a kind of neural network with two or more identical subnetworks that have the same architecture and share the same parameters and weights \cite{lecun1994siamese}. 
By deriving fixed-sized embeddings and computing similarities, siamese network systems have proven effective in modeling the relationship between two text sequences \cite{conneau2017sialstm, yang2018siatransformers, Reimers2019SentenceBERTSE}.

% which measures the semantic relatedness of natural language query and code. 
% Siamese network is a special neural network that uses two or more identical subnetworks to 
% that have the same architecture and share the same parameters and weights \cite{lecun1994siamese}, which are typically used in tasks that require 

% \paragraph{CodeBERT}
We use a pretrained CodeBERT \cite{Feng2020CodeBERTAP} as the encoder to map any text sequence to a $d$-dimensional real-valued vectors. 
CodeBERT is a bimodal model for natural language and programming language which enables high-quality text and code embeddings to be derived. 
Specifically, it shares exactly the same architecture as RoBERTa \cite{Liu2019RoBERTaAR}, which is a bidirectional Transformer with 12 layers, 768 dimensional hidden states, and 12 attention heads,  and is repretrained by masked language modeling and replaced token detection objectives on CodeSearchNet corpus \cite{husain2019codesearchnet}. 
% It is a bimodal model for programming language (PL) and natural language (NL), and is pretrained by masked language modeling and replaced token detection objectives on the CodeSearchNet corpus. 
% The model supports both unimodal and bimodal sequence inputs with a class token $[CLS]$ in front of the sequence and a seperation token $[SEP]$ to split two modalities of data. 

% \subsubsection*{Siamese CodeBERT}
% \paragraph{Siamese CodeBERT}
% Our Code Contrastive Learning (CoCLR) method uses a siamese CodeBERT structure to compute the semantic relatedness of natural language query and code. Siamese network uses two or more identical subnetworks that have the same architecture and share the same parameters and weights \cite{lecun1994siamese}, which are typically used in tasks that require modeling the relationship between two sequences \cite{Reimers2019SentenceBERTSE}.
% \paragraph{Encoder}
For each query $q_i$ and code $c_i$, we concatenate a $[CLS]$ token in front of the sequence 
and a $[SEP]$ token at the end. 
Then we feed the query and code sequences into the CodeBERT encoder to obtain contextualized embeddings, respectively. Here we use the pooled output of $[CLS]$ token as the representations: 
\begin{equation}
\small
    \mathbf{q}_i = \mathbf{CodeBERT}(q_i), \quad \mathbf{c}_i = \mathbf{CodeBERT}(c_i).
\end{equation}
  
% Given a pair of query $q_i$ and code $c_i$, siamese CodeBERT first encodes query and code with CodeBERT $f(\cdot)$ respectively. Here we use the pooled output of $[CLS]$ token as the representations: 
% \begin{equation}
%     \mathbf{q}_i = f(q_i), \quad \mathbf{c}^i = f(c_i).
% \end{equation}

Next we perform query-code matching through a multi-layer perceptron. Following \citet{Chen2017ESIM} and \citet{Mou2016NLITree}, we concatenate the query embedding $\mathbf{q}_i$ and code embedding $\mathbf{c}_i$ with the element-wise difference $\mathbf{q}_i-\mathbf{c}_i$ and element-wise product $\mathbf{q}_i \bigodot \mathbf{c}_i$, followed by a 1-layer feed-forward neural network, to obtain a relation embedding:
\begin{equation}
\small
    \mathbf{r}^{(i,i)} = \tanh (\mathbf{W}_{1} \cdot [\mathbf{q}_i, \mathbf{c}_i, \mathbf{q}_i-\mathbf{c}_i, \mathbf{q}_i \bigodot \mathbf{c}_i]).
\end{equation}
We expect such an operation can help sharpen the cross information between query and code to capture better matching relationships such as contradiction. 

Then we put the relation embedding $\mathbf{r}^{(i,i)}$ into a final 1-layer perceptron classifier with a sigmoid output layer: $s^{(i,i)} = sigmoid (\mathbf{W}_2 \cdot \mathbf{r}^{(i,i)})$. Score $s^{(i, i)}$ can be viewed as the similarity of query $q_i$ and code $c_i$.

% \subsubsection*{Vanilla Loss}
% \paragraph{Vanilla Loss}
To train the base siamese network, we use a binary cross entropy loss as the objective function: 
\begin{equation}\label{eq:b}
\small
    \mathcal{L}_{b} = - [y_i \cdot \log s^{(i,i)} + (1-y_i) \log (1-s^{(i,i)})],
\end{equation}
where $y_i$ is the label of $(q_i, c_i)$.

\subsection{Code Contrastive Learning}
% Inspired by SimCLR \cite{chen2020simclr}, CoCLR learns to maximize agreement between positive query-code pairs via a contrastive loss in the latent space. 
Now we incorporate code contrastive learning into the siamese network with CodeBERT. Contrastive learning aims to learn representations by enforcing similar objects to be closer while keeping dissimilar objects further apart. 
It is often accompanied with leveraging task-specific inductive bias to augment similar and dissimilar examples. 
In this work, given an example of query and code $(q_i, c_i)$, we define our contrastive learning task on example itself, in-batch augmented examples $(q_i, c_j)$, and augmented example with rewritten query $(q_i^{\prime}, c_i)$.
Hence, the overall training objective can be formulated as:

\begin{equation}
\small
    \mathcal{L} = \mathcal{L}_{b} + \mathcal{L}_{ib} + \mathcal{L}_{qr}.
\end{equation}

% \subsubsection*{Negative Examples} 
\paragraph{In-Batch Augmentation (IBA)}
A straightforward augmentation method
% for siamese networks
is to use in-batch data, where a query and a randomly sampled code are considered as dissimilar and forced away by the models.
% A straightforward way is to use in-batch data for augmentation, where we treat a query and a randomly sampled code as dissimilar and enforcing them away.
% The negative examples come from mini-batch data during training. 
Specifically, we randomly sample $n$ examples $\{(q_1, c_1), (q_2, c_2), \dots, (q_n, c_n)\}$ from a mini-batch. For $(q_i, c_i)$, we pair query $q_i$ with the other $N - 1$ codes within the mini-batch and treat the $N-1$ pairs as dissimilar. 
Let $s^{(i,j)}$ denote the similarity of query $q_i$ and code $c_j$, the loss function of the example with IBA is defined as:
\begin{equation}\label{eq:ib}
\small
    \mathcal{L}_{ib} = - \frac{1}{n-1} \sum_{\mbox{\tiny$\begin{array}{c}
j=1\\j\neq i\end{array}$}}^{n} \log(1-s^{(i,j)}) ,
\end{equation}

% \subsubsection*{Positive Examples}
\paragraph{Query-Rewritten Augmentation (QRA)}
The in-batch augmentation only creates dissimilar pairs from the mini-batch, 
which ignores to augment similar pairs for learning positive relations. 
To remedy this, we propose to augment positive examples by rewriting queries.
Inspired by the feature that web queries are often brief and not necessarily grammatically correct, we assume that the rewritten query with minor modifications shares the same semantics as the original one. Therefore, an augmented pair with a rewritten query from a positive pair can also be treated as positive. 
% As a unique characteristic, web queries are sometimes brief and not necessarily grammatically correct. Inspired by this, we assume that the rewritten query with minor modifications shares the same semantics with the original one. 

% The pseudo positive examples in CoCLR are created explicitly. 
% Inspired by the characteristic that web queries are often brief and not necessarily grammatically correct, we assume that the augmented query with minor modifications shares the same semantics with the original one.
% So we apply a data augmentor to create pseudo web queries for code. 

\begin{table*}[htbp]
\small
  \centering
    \begin{tabular}{llcc}
    \toprule
    Model & Data & Code Question Answering & Code Search \\
    \midrule 
    RoBERTa\footnotemark[2]  & CSN   & 40.34 &  0.18 \\
    CodeBERT\footnotemark[2]  & CSN   & 47.80 &  51.29 \\
    CodeBERT & CSN + CoSQA &  52.87 &  54.41 \\
    CodeBERT + CoCLR & CSN + CoSQA & \textbf{63.38} &  \textbf{64.66} \\
    \bottomrule
    \end{tabular}%
  \caption{Evaluation on code question answering and code search. CSN denotes CodeSearchNet Python corpus. By incorporating CoCLR method, siamese network with CodeBERT outperforms the existing baseline approaches.}
  \label{tab:overall-result}%
\end{table*}%

Specifically, given a pair of query $q_i$ and code $c_i$ with $y_i=1$, we rewrite $q_i$ into $q_i^{\prime}$ in one of the three ways: randomly \textit{deleting} a word, randomly \textit{switching} the position of two words, and randomly \textit{copying} a word. As shown in Section \ref{sec:data-augementor-ablation}, switching position best helps increase the performance. 

For any augmented positive examples, we also apply IBA on them.  
% both $\mathcal{L}_b$ and $\mathcal{L}_i$ on $(q_i^{\prime}, c_i)$.
Therefore the loss function for the example with QRA is:
\begin{equation}
\small
    \mathcal{L}_{qr} = \mathcal{L}_b^{\prime} + \mathcal{L}_{ib}^{\prime},
    % \mathcal{L}_q = y_i \cdot (\mathcal{L}_b^{\prime} + \mathcal{L}_n^{\prime}).
\end{equation}
where $\mathcal{L}_b^{\prime}$ and $\mathcal{L}_{ib}^{\prime}$ can be obtained by Eq. \ref{eq:b} and Eq. \ref{eq:ib} by only change $q_i$ to $q_i^{\prime}$.
% % 应该补一个实验：p‘和p的sample的负例一样和不一样的比较

\section{Experiments}
% In this section, we evaluate our CoCLR method on query-code matching. 
We experiment on two tasks, including code question answering and natural language code search. 
We report model comparisons and give detailed analyses from different perspectives.

% Then we perform an ablation study of the different aspects of CoCLR in order to get a better understanding of their relative importance. In addition, we give a thorough analysis of the effects of different code component in code search.

\subsection{Experiment Settings} 

% \paragraph{Dataset}
We train the models on the CoSQA dataset and evaluate them on two tasks: code question answering and code search.

On code question answering, we randomly split CoSQA into 20,000 training and 604 validation examples. As for the test set, we directly use the WebQueryTest in CodeXGLUE benchmark, which is a testing set of Python code question answering with 1,046 query-code pairs and their expert annotations. 

% For code search task, we randomly select 500 queries with positive code answers to form both the validation and test sets. 
On code search, we randomly divide the CoSQA into training, validation, and test sets in the number of 19604:500:500, and restrict the instances for validation and testing are all positive. We fix a code database with 6,267 different codes in CoSQA.

\paragraph{Baseline Methods}
CoSQA is a new dataset, and there are no previous models designed specifically for it. Hence, we simply choose RoBERTa-base \cite{Liu2019RoBERTaAR} and CodeBERT \cite{Feng2020CodeBERTAP} as the baseline methods. The baseline methods are trained on CodeSearchNet Python corpus with balanced positive examples. Negative samples consist of a balanced number of instances with randomly replaced code.

\paragraph{Evaluation Metric}
We use accuracy as the evaluation metric on code question answering and Mean Reciprocal Rank (MRR) on code search.
% On code question answering, we use accuracy as the metric. On code search, we use Mean Reciprocal Rank (MRR) as the metric. 

\paragraph{Implementation Details}
% 模型的一些参数
We initialize CoCLR with \textit{microsoft/codebert-base}\footnote{ \href{https://github.com/microsoft/CodeBERT}{https://github.com/microsoft/CodeBERT}} repretrained on CodeSearchNet Python Corpus \cite{husain2019codesearchnet}. We use the AdamW optimizer \cite{Loshchilov2019DecoupledWD} and set the batch size to 32 on the two tasks. On code question answering, we set the learning rate to 1e-5, warm-up rate to 0.1. On code search, we set the learning rate to 1e-6. All hyper-parameters are tuned to the best on the validation set. 
All experiments are performed on an  NVIDIA Tesla V100 GPU with 16GB memory.

% \subsection{Overall Results}
\subsection{Model Comparisons}
Table \ref{tab:overall-result} shows the experimental results on the tasks of code question answering and code search. We can observe that:

(1) By leveraging the CoSQA dataset, siamese network with CodeBERT achieves overall performance enhancement on two tasks, especially for CodeXGLUE WebQueryTest, which is an open challenge but without direct training data. 
% This result demonstrates that CoSQA dataset is of high-quality and can be leveraged to train  
The result demonstrates the high-quality of CoSQA and its potential to be the training set of WebQueryTest.

(2) By integrating the code contrastive learning method, siamese network with CodeBERT further achieves significant performance gain on both tasks. Especially on the task of WebQueryTest, CoCLR achieves the new state-of-the-art result by increasing 15.6\%, which shows the effectiveness of our proposed approach.

% \section{Analysis}
% In this section, we perform comprehensive analyses to investigate the factors that influence the retrieval results of CoCLR, aiming to reveal the characteristics of CoCLR and query-code matching.

% \begin{table}[t]
% \small
%   \centering
%     \begin{tabular}{lc}
%     \toprule
%     Data & MRR \\
%     \midrule
%     original CoSQA only  & 54.41 \\
%     \quad + positive (delete) & 55.24 \\
%     \quad + positive (copy) & 54.82 \\
%     \quad + positive (switch) & 55.66 \\
%     \quad + negative & 63.51 \\
%     \quad + negative + positive (delete) & 63.41 \\
%     \quad + negative + positive (copy) & 63.97 \\
%     \quad + negative + positive (switch) & \textbf{64.66} \\
%     \bottomrule
%     \end{tabular}%
%   \caption{Performance of CodeBERT with CoCLR component on code search task.}
%   \label{tab:ablation-constrastive-result}%
% \end{table}%
\begin{table}[t]
\small
  \centering
    \begin{tabular}{lc}
    \toprule
    Augmentations & MRR \\
    \midrule
    no augmentations  & 54.41 \\
    \quad + query-rewritten (delete) & 55.24 \\
    \quad + query-rewritten (copy) & 54.82 \\
    \quad + query-rewritten (switch) & 55.66 \\
    \quad + in-batch & 63.51 \\
    \quad + in-batch + query-rewritten (delete) & 63.41 \\
    \quad + in-batch + query-rewritten (copy) & 63.97 \\
    \quad + in-batch + query-rewritten (switch) & \textbf{64.66} \\
    \bottomrule
    \end{tabular}%
  \caption{Performance of CodeBERT with different augmentations in CoCLR on code search.}
  \label{tab:ablation-constrastive-result}%
\end{table}%

% \subsection{Ablation Study}

\subsection{Analysis: Effects of CoCLR}\label{sec:data-augementor-ablation}
% \subsection{Ablation Study}\label{sec:data-augementor-ablation}
To investigate the effects of CoCLR in query-code matching, we perform ablation study to analyze the major components in our contrastive loss that are of importance to help achieve good performance. We conduct experiments on the CoSQA code search task, using the following settings: (i) fine-tuning with vanilla binary cross-entropy loss only, (ii) fine-tuning with additional in-batch augmentation (IBA) loss, (iii) fine-tuning with additional query-rewritten augmentation (QRA) loss, (vi) fine-tuning with both additional IBA and QRA loss. And for QRA loss, we also test the three rewriting methods when applied individually. The results are listed in Table \ref{tab:ablation-constrastive-result}. We can find that: 

(1) Both incorporating IBA and QRA individually or together improve models' performance. This indicates the advantage of applying code contrastive learning for code search.

% (2) Among the three positive data augmentation approaches, \textit{switch} method consistently improves the performance most. 
(2) No matter integrating IBA or not, the model with QRA by \textit{switching} method performs better than models with the other two methods.
We attribute the phenomenon to the fact that web queries do not necessarily have accurate grammar. So switching the positions of two words in the query better maximizes the agreement between the positive example and the pseudo positive example than the other two augmentations, which augments better examples to learn representations.

(3) Comparing the two augmentations, adding IBA achieves more performance gain than QRA (1.25\% versus 9.10\%). 
As the numbers of examples with QRA and examples with IBA are not equal under two settings, we further evaluate the model with only one more example with IBA. The MRR is 55.52\%, which is comparable to the performance of adding one more example with QRA. 
This suggests that there may be no difference between adding examples with IBA or examples with QRA. Instead, the number of high-quality examples is important for training. Similar findings are also reported in \citet{Sun2020ContrastiveDO}, and a theoretical analysis is provided in \citet{Arora2019ATA}. 
% This suggests that there may not be absolute better or worse for negative examples and positive examples. Instead, adding more examples to training does matter. 

% \footnotetext[6]{https://github.com/microsoft/CodeXGLUE}

\subsection{Analysis: Effects of Code Components}\label{sec:code-component}
To explore the effects of different components of code in query-code matching, we evaluate CoCLR on code search and process the codebase by the following operations: (i) removing the function header, (ii) removing the natural language documentation, (iii) removing the code statements in the function body. We also combine two of the above operations to see the performance.  From the results exhibited in Table \ref{tab:ablation-code-component-result}, we can find that: by removing code component, the result of removing  documentation drops more than those of removing header and removing function body. This demonstrates the importance of natural language documentation in code search. Since documentation shares the same modality with the query and briefly describes the functionality of the code, it may be more semantically related to the query. Besides, it also reveals the importance of using web queries rather than treating documentation as queries in code search datasets, which liberates models from the matching between documentation with code to the matching between query with documentation and code. 

\begin{table}[t]
\small
  \centering
    \begin{tabular}{lc}
    \toprule
    Code Component & MRR \\
    \midrule
    complete code & \textbf{64.66} \\
    \quad w/o header & 62.01 \\
    \quad w/o body & 59.11 \\
    \quad w/o documentation & 58.54 \\
    \quad w/o header \& body & 52.89 \\
    \quad w/o header \& documentation & 43.35 \\
    \quad w/o body \& documentation & 42.71 \\
    \bottomrule
    \end{tabular}%
  \caption{Performance of CoCLR-incorporated CodeBERT trained and tested with different code components on code search.}
  \label{tab:ablation-code-component-result}%
\end{table}%

\section{Conclusion}

In this paper, we focus on the matching problem of the web query and code. We develop a large-scale human-annotated query-code matching dataset CoSQA, which contains 20,604 pairs of real-world web queries and Python functions with documentation. We demonstrate that CoSQA is an ideal dataset for code question answering and code search. We also propose a novel code contrastive learning method, named CoCLR, to incorporate artificially generated instances into training. We find that model with CoCLR outperforms the baseline models on code search and code question answering tasks. 
We perform detailed analysis to 
investigate the effects of CoCLR components and code components in query-code matching. 
We believe our annotated CoSQA dataset will be useful for 
other tasks that involve aligned text and code, such as code summarization and code synthesis.

\section*{Acknowledgement}
We thank all anonymous reviewers for their useful comments. We also thank Zenan Xu, Daya Guo, Shuai Lu, Wanjun Zhong and Siyuan Wang for valuable discussions and feedback during the paper writing process.

% \clearpage

\bibliography{anthology,acl2021}

\begin{thebibliography}{39}
\expandafter\ifx\csname natexlab\endcsname\relax\def\natexlab#1{#1}\fi

\bibitem[{Allamanis et~al.(2018)Allamanis, Barr, Devanbu, and
  Sutton}]{miltos2018survey}
Miltiadis Allamanis, Earl~T. Barr, Premkumar Devanbu, and Charles Sutton. 2018.
\newblock A survey of machine learning for big code and naturalness.
\newblock \emph{ACM Computing Survey}.

\bibitem[{Arora et~al.(2019)Arora, Khandeparkar, Khodak, Plevrakis, and
  Saunshi}]{Arora2019ATA}
S.~Arora, Hrishikesh Khandeparkar, M.~Khodak, Orestis Plevrakis, and Nikunj
  Saunshi. 2019.
\newblock A theoretical analysis of contrastive unsupervised representation
  learning.
\newblock In \emph{ICML}.

\bibitem[{Bajracharya et~al.(2006)Bajracharya, Ngo, Linstead, Dou, Rigor,
  Baldi, and Lopes}]{Bajracharya2006SourcererAS}
S.~Bajracharya, Trung~Chi Ngo, Erik Linstead, Yimeng Dou, Paul Rigor, P.~Baldi,
  and C.~Lopes. 2006.
\newblock Sourcerer: a search engine for open source code supporting
  structure-based search.
\newblock In \emph{OOPSLA '06}.

\bibitem[{Bromley et~al.(1994)Bromley, Guyon, LeCun, S\"{a}ckinger, and
  Shah}]{lecun1994siamese}
Jane Bromley, Isabelle Guyon, Yann LeCun, Eduard S\"{a}ckinger, and Roopak
  Shah. 1994.
\newblock Signature verification using a "siamese" time delay neural network.
\newblock In \emph{Advances in Neural Information Processing Systems}.

\bibitem[{Cambronero et~al.(2019)Cambronero, Li, Kim, Sen, and
  Chandra}]{Cambronero2019WhenDL}
Jos{\'e} Cambronero, Hongyu Li, S.~Kim, K.~Sen, and S.~Chandra. 2019.
\newblock When deep learning met code search.
\newblock \emph{Proceedings of the 2019 27th ACM Joint Meeting on European
  Software Engineering Conference and Symposium on the Foundations of Software
  Engineering}.

\bibitem[{Chen et~al.(2017)Chen, Zhu, Ling, Wei, Jiang, and
  Inkpen}]{Chen2017ESIM}
Qian Chen, Xiao-Dan Zhu, Zhenhua Ling, Si~Wei, Hui Jiang, and D.~Inkpen. 2017.
\newblock Enhanced lstm for natural language inference.
\newblock In \emph{ACL}.

\bibitem[{Conneau et~al.(2017)Conneau, Kiela, Schwenk, Barrault, and
  Bordes}]{conneau2017sialstm}
Alexis Conneau, Douwe Kiela, Holger Schwenk, Lo{\"\i}c Barrault, and Antoine
  Bordes. 2017.
\newblock Supervised learning of universal sentence representations from
  natural language inference data.
\newblock In \emph{EMNLP}.

\bibitem[{Feng et~al.(2020)Feng, Guo, Tang, Duan, Feng, Gong, Shou, Qin, Liu,
  Jiang, and Zhou}]{Feng2020CodeBERTAP}
Zhangyin Feng, Daya Guo, Duyu Tang, N.~Duan, X.~Feng, Ming Gong, Linjun Shou,
  B.~Qin, Ting Liu, Daxin Jiang, and M.~Zhou. 2020.
\newblock Codebert: A pre-trained model for programming and natural languages.
\newblock In \emph{Findings of the Association for Computational Linguistics:
  EMNLP 2020}.

\bibitem[{Gu et~al.(2018)Gu, Zhang, and Kim}]{gu2018deepcs}
Xiaodong Gu, Hongyu Zhang, and Sunghun Kim. 2018.
\newblock Deep code search.
\newblock In \emph{Proceedings of ICSE}.

\bibitem[{Guo et~al.(2020)Guo, Ren, Lu, Feng, Tang, Liu, Zhou, Duan, Yin,
  Jiang, and Zhou}]{Guo2020GraphCodeBERTPC}
Daya Guo, Shuo Ren, Shuai Lu, Zhangyin Feng, Duyu Tang, Shujie Liu, L.~Zhou,
  N.~Duan, Jian Yin, Daxin Jiang, and M.~Zhou. 2020.
\newblock Graphcodebert: Pre-training code representations with data flow.
\newblock In \emph{ICLR}.

\bibitem[{Haldar et~al.(2020)Haldar, Wu, Xiong, and
  Hockenmaier}]{haldar2020multi}
Rajarshi Haldar, Lingfei Wu, JinJun Xiong, and Julia Hockenmaier. 2020.
\newblock A multi-perspective architecture for semantic code search.
\newblock In \emph{Proceedings of ACL}.

\bibitem[{Heyman and Cutsem(2020)}]{Heyman2020SODS}
Geert Heyman and Tom~Van Cutsem. 2020.
\newblock Neural code search revisited: Enhancing code snippet retrieval
  through natural language intent.
\newblock \emph{ArXiv}, abs/2008.12193.

\bibitem[{Husain et~al.(2019)Husain, Wu, Gazit, Allamanis, and
  Brockschmidt}]{husain2019codesearchnet}
Hamel Husain, Ho-Hsiang Wu, Tiferet Gazit, Miltiadis Allamanis, and Marc
  Brockschmidt. 2019.
\newblock {CodeSearchNet} challenge: Evaluating the state of semantic code
  search.
\newblock \emph{arXiv preprint arXiv:1909.09436}.

\bibitem[{Krippendorff(1980)}]{Krippendorff1980KrippendorffKC}
K.~Krippendorff. 1980.
\newblock Krippendorff, klaus, content analysis: An introduction to its
  methodology . beverly hills, ca: Sage, 1980.

\bibitem[{Li et~al.(2019)Li, Kim, and Chandra}]{Li2019NeuralCS}
Hongyu Li, S.~Kim, and S.~Chandra. 2019.
\newblock Neural code search evaluation dataset.
\newblock \emph{ArXiv}, abs/1908.09804.

\bibitem[{Liu et~al.(2020{\natexlab{a}})Liu, Xia, Lo, Gao, Yang, and
  Grundy}]{Liu2020OpportunitiesAC}
C.~Liu, Xin Xia, David Lo, Cuiyun Gao, Xiaohu Yang, and J.~Grundy.
  2020{\natexlab{a}}.
\newblock Opportunities and challenges in code search tools.
\newblock \emph{ArXiv}, abs/2011.02297.

\bibitem[{Liu et~al.(2020{\natexlab{b}})Liu, Xia, Lo, Liu, Hassan, and
  Li}]{Liu2020SimplifyingDM}
Chao Liu, Xin Xia, David Lo, Zhiwei Liu, A.~Hassan, and Shanping Li.
  2020{\natexlab{b}}.
\newblock Simplifying deep-learning-based model for code search.
\newblock \emph{ArXiv}, abs/2005.14373.

\bibitem[{Liu et~al.(2019{\natexlab{a}})Liu, Kim, Murali, Chaudhuri, and
  Chandra}]{liu2019queryexpand}
Jason Liu, Seohyun Kim, Vijayaraghavan Murali, Swarat Chaudhuri, and Satish
  Chandra. 2019{\natexlab{a}}.
\newblock Neural query expansion for code search.
\newblock In \emph{Proceedings of the 3rd ACM SIGPLAN International Workshop on
  Machine Learning and Programming Languages}. ACM.

\bibitem[{Liu et~al.(2019{\natexlab{b}})Liu, Ott, Goyal, Du, Joshi, Chen, Levy,
  Lewis, Zettlemoyer, and Stoyanov}]{Liu2019RoBERTaAR}
Y.~Liu, Myle Ott, Naman Goyal, Jingfei Du, Mandar Joshi, Danqi Chen, Omer Levy,
  M.~Lewis, Luke Zettlemoyer, and Veselin Stoyanov. 2019{\natexlab{b}}.
\newblock Roberta: A robustly optimized bert pretraining approach.
\newblock \emph{ArXiv}, abs/1907.11692.

\bibitem[{Loshchilov and Hutter(2019)}]{Loshchilov2019DecoupledWD}
I.~Loshchilov and F.~Hutter. 2019.
\newblock Decoupled weight decay regularization.
\newblock In \emph{ICLR}.

\bibitem[{Lu et~al.(2015)Lu, Sun, Wang, Lo, and Duan}]{Lu2015QueryEV}
Meili Lu, Xiaobing Sun, S.~Wang, D.~Lo, and Yucong Duan. 2015.
\newblock Query expansion via wordnet for effective code search.
\newblock \emph{2015 IEEE 22nd International Conference on Software Analysis,
  Evolution, and Reengineering (SANER)}, pages 545--549.

\bibitem[{Lu et~al.(2021)Lu, Guo, Ren, Huang, Svyatkovskiy, Blanco, Clement,
  Drain, Jiang, Tang, Li, Zhou, Shou, Zhou, Tufano, Gong, Zhou, Duan,
  Sundaresan, Deng, Fu, and Liu}]{codexglue}
Shuai Lu, Daya Guo, Shuo Ren, Junjie Huang, Alexey Svyatkovskiy, Ambrosio
  Blanco, Colin Clement, Dawn Drain, Daxin Jiang, Duyu Tang, Ge~Li, L.~Zhou,
  Linjun Shou, Long Zhou, Michele Tufano, Ming Gong, Ming Zhou, Nan Duan, Neel
  Sundaresan, Shao~Kun Deng, Shengyu Fu, and Shujie Liu. 2021.
\newblock Codexglue: A machine learning benchmark dataset for code
  understanding and generation.
\newblock \emph{ArXiv}, abs/2102.04664.

\bibitem[{Lv et~al.(2015)Lv, Zhang, Lou, Wang, Zhang, and
  Zhao}]{Lv2015CodeHowEC}
Fei Lv, H.~Zhang, Jian-Guang Lou, S.~Wang, D.~Zhang, and Jianjun Zhao. 2015.
\newblock Codehow: Effective code search based on api understanding and
  extended boolean model (e).
\newblock \emph{2015 30th IEEE/ACM International Conference on Automated
  Software Engineering (ASE)}, pages 260--270.

\bibitem[{Miceli~Barone and Sennrich(2017)}]{sennrich2017codedoc}
Antonio~Valerio Miceli~Barone and Rico Sennrich. 2017.
\newblock A parallel corpus of python functions and documentation strings for
  automated code documentation and code generation.
\newblock In \emph{Proceedings of IJCNLP}.

\bibitem[{Mou et~al.(2016)Mou, Men, Li, Xu, Zhang, Yan, and
  Jin}]{Mou2016NLITree}
Lili Mou, Rui Men, Ge~Li, Yan Xu, Lu~Zhang, Rui Yan, and Zhi Jin. 2016.
\newblock Natural language inference by tree-based convolution and heuristic
  matching.
\newblock In \emph{ACL}.

\bibitem[{Nie et~al.(2016)Nie, Jiang, Ren, Sun, and Li}]{Nie2016QueryEB}
Liming Nie, He~Jiang, Zhilei Ren, Zeyi Sun, and Xiaochen Li. 2016.
\newblock Query expansion based on crowd knowledge for code search.
\newblock \emph{IEEE Transactions on Services Computing}, 9:771--783.

\bibitem[{Rahman(2019)}]{Rahman2019SupportingCS}
M.~M. Rahman. 2019.
\newblock Supporting code search with context-aware, analytics-driven,
  effective query reformulation.
\newblock \emph{2019 IEEE/ACM 41st International Conference on Software
  Engineering: Companion Proceedings (ICSE-Companion)}, pages 226--229.

\bibitem[{Rahman et~al.(2019)Rahman, Roy, and Lo}]{Rahman2019AutomaticQR}
M.~M. Rahman, C.~Roy, and D.~Lo. 2019.
\newblock Automatic query reformulation for code search using crowdsourced
  knowledge.
\newblock \emph{Empirical Software Engineering}, pages 1--56.

\bibitem[{Reimers and Gurevych(2019)}]{Reimers2019SentenceBERTSE}
Nils Reimers and Iryna Gurevych. 2019.
\newblock Sentence-bert: Sentence embeddings using siamese bert-networks.
\newblock In \emph{EMNLP/IJCNLP}.

\bibitem[{Sachdev et~al.(2018)Sachdev, Li, Luan, Kim, Sen, and
  Chandra}]{Sachdev2018RetrievalOS}
Saksham Sachdev, H.~Li, Sifei Luan, S.~Kim, K.~Sen, and S.~Chandra. 2018.
\newblock Retrieval on source code: a neural code search.
\newblock \emph{Proceedings of the 2nd ACM SIGPLAN International Workshop on
  Machine Learning and Programming Languages}.

\bibitem[{Sun et~al.(2020)Sun, Gan, Cheng, Fang, Wang, and jing
  Liu}]{Sun2020ContrastiveDO}
S.~Sun, Zhe Gan, Y.~Cheng, Yuwei Fang, Shuohang Wang, and Jing jing Liu. 2020.
\newblock Contrastive distillation on intermediate representations for language
  model compression.
\newblock In \emph{EMNLP}.

\bibitem[{Wan et~al.(2019)Wan, Shu, Sui, Xu, Zhao, Wu, and
  Yu}]{yao2019mulmodalatt}
Yao Wan, Jingdong Shu, Yulei Sui, Guandong Xu, Zhou Zhao, Jian Wu, and
  Philip~S. Yu. 2019.
\newblock Multi-modal attention network learning for semantic source code
  retrieval.
\newblock \emph{2019 34th IEEE/ACM International Conference on Automated
  Software Engineering (ASE)}, pages 13--25.

\bibitem[{Yan et~al.(2020)Yan, Yu, Chen, Shen, and Jiang}]{yan2020benchcs}
Shuhan Yan, Hang Yu, Yuting Chen, Beijun Shen, and Lingxiao Jiang. 2020.
\newblock Are the code snippets what we are searching for? a benchmark and an
  empirical study on code search with natural-language queries.
\newblock In \emph{Proceedings of SANER}.

\bibitem[{Yang et~al.(2018)Yang, Yuan, Cer, Kong, Constant, Pilar, Ge, Sung,
  Strope, and Kurzweil}]{yang2018siatransformers}
Yinfei Yang, Steve Yuan, Daniel Cer, Sheng-yi Kong, Noah Constant, Petr Pilar,
  Heming Ge, Yun-Hsuan Sung, Brian Strope, and Ray Kurzweil. 2018.
\newblock Learning semantic textual similarity from conversations.
\newblock In \emph{Proceedings of The Third Workshop on Representation Learning
  for {NLP}}.

\bibitem[{Yao et~al.(2019)Yao, Peddamail, and Sun}]{yao2019coacor}
Ziyu Yao, Jayavardhan~Reddy Peddamail, and Huan Sun. 2019.
\newblock Coacor: Code annotation for code retrieval with reinforcement
  learning.
\newblock In \emph{Proceedings of WWW}.

\bibitem[{Yao et~al.(2018)Yao, Weld, Chen, and Sun}]{yao2018staqc}
Ziyu Yao, Daniel~S Weld, Wei-Peng Chen, and Huan Sun. 2018.
\newblock {S}ta{QC}: A systematically mined question-code dataset from stack
  overflow.
\newblock In \emph{Proceedings of WWW}.

\bibitem[{Ye et~al.(2016)Ye, Shen, Ma, Bunescu, and Liu}]{Ye2016FromWE}
Xin Ye, Hui Shen, Xiao Ma, Razvan~C. Bunescu, and Chang Liu. 2016.
\newblock From word embeddings to document similarities for improved
  information retrieval in software engineering.
\newblock \emph{2016 IEEE/ACM 38th International Conference on Software
  Engineering (ICSE)}.

\bibitem[{Yin et~al.(2018)Yin, Deng, Chen, Vasilescu, and
  Neubig}]{yin2018mining}
Pengcheng Yin, Bowen Deng, Edgar Chen, Bogdan Vasilescu, and Graham Neubig.
  2018.
\newblock Learning to mine aligned code and natural language pairs from stack
  overflow.
\newblock In \emph{International Conference on Mining Software Repositories}.

\bibitem[{Zhao and Sun(2020)}]{zhao2020adversarialcodesearch}
Jie Zhao and Huan Sun. 2020.
\newblock Adversarial training for code retrieval with question-description
  relevance regularization.
\newblock In \emph{Findings of the Association for Computational Linguistics:
  EMNLP 2020}.

\end{thebibliography}
\bibliographystyle{acl_natbib}

% \clearpage

\appendix

\section{Heuristics for Query Filtering}
\label{appen:query-filter-rules}
In this section, we introduce our heuristic rules to filter potential queries without code search intent. Basically, the rules are created from keyword templates and we follow the six categories of queries without code search intent to derive the keywords. Note that vague queries are morphologically variable so we ignore this categories. The keywords are shown in Table \ref{tab:appendix-keywords}.

\begin{table}[htbp]
\small
  \centering
%   \resizebox{1.0\textwidth}{!}{
    % \begin{tabular}{p{3cm}p{8cm}p{2cm}}
    \begin{tabular}{m{1.5cm}<{\centering}m{5.4cm}}
    
    \toprule
    Categories & Keywords   \\
    \midrule
    Debugging & 
            exception, index out of, ignore, omit, stderr, try \dots except, debug, no such file or directory, warning, 
        \\ \\
    Conceptual & 
            vs, versus, difference, advantage, benefit, drawback,
            interpret, understand, 
            cannot, can’t, couldn’t, could not, 
            how many, how much, too much, too many, more, less,
            what if, what happens, what is, what are,
            when, where, which, why, reason,
            how do \dots work, how \dots works, how does \dots work,
            need, require, wait, turn \dots on/off, turning \dots on/off, 
        \\ \\
    Programming Knowledge & 
            tutorial, advice, course, proposal, discuss, suggestion,
            parameter, argument, statement, class, import,
            inherit, operator, override, decorator, descriptor,
            declare, declaration
        \\ \\
    Tools Usage & 
            console, terminal, open python, studio, ide,
            ipython, jupyter, notepad, notebook, vim,
            pycharm, vscode, eclipse, sublime, emacs, utm,
            komodo, pyscripter, eric, c\#, access control,
            pip, install, library, module, launch, version,
            ip address, ipv, get \dots ip, check \dots ip, valid \dots ip,
        \\ \\
    Others & 
          unicode, python command, 
            ``()", ``.", ``\_", ``:", ``@", ``=", ``\textgreater", ``\textless", ``-"
        \\
    
    \bottomrule
    \end{tabular}%
%   }
  \caption{Keywords of queries without code search intent in five categories.} 
  \label{tab:appendix-keywords}%
\end{table}%

\end{document}